\documentclass[a4paper,twoside]{article}

\usepackage{epsfig}
\usepackage{subcaption}
\usepackage{calc}
\usepackage{amssymb}
\usepackage{amstext}
\usepackage{amsmath}
\usepackage{amsthm}
\usepackage{multicol}
\usepackage{pslatex}
\usepackage{apalike}
\usepackage[bottom]{footmisc}

\usepackage{bbm}
\usepackage{hyperref}

\usepackage{SCITEPRESS}     

\begin{document}

\title{Surface-biased Multi-Level Context 3D Object Detection}

\author{\authorname{Sultan Abu Ghazal\sup{1}\orcidAuthor{0000-0003-0416-5964}, Jean Lahoud\sup{1}\orcidAuthor{0000-0003-0315-6484} and Rao Anwer\sup{1}\orcidAuthor{0000-0002-9041-2214}}
\affiliation{\sup{1}Mohamed Bin Zayed University of Artificial Intelligence, Abu Dhabi, UAE}
\email{\{sultan.abughazal, jean.lahoud, rao.anwer\}@mbzuai.ac.ae}
}

\keywords{Detection, Segmentation, Multi-Level Context, 3D.}

\abstract{
Object detection in 3D point clouds is a crucial task in a range of computer vision applications including robotics, autonomous cars, and augmented reality. This work addresses the object detection task in 3D point clouds using a highly efficient, surface-biased, feature extraction method \cite{wang2022rbgnet}, that also captures contextual cues on multiple levels. We propose a 3D object detector that extracts accurate feature representations of object candidates and leverages self-attention on point patches, object candidates, and on the global scene in 3D scene. Self-attention is proven to be effective in encoding correlation information in 3D point clouds by \cite{xie2020mlcvnet}. While other 3D detectors focus on enhancing point cloud feature extraction by selectively obtaining more meaningful local features \cite{wang2022rbgnet} where contextual information is overlooked. To this end, the proposed architecture uses ray-based surface-biased feature extraction and multi-level context encoding to out perform the state-of-the-art 3D object detector. In this work, 3D detection experiments are performed on scenes from the ScanNet dataset whereby the self-attention modules are introduced one after the other to isolate the effect of self-attention at each level. The code is available at \href{https://github.com/SultanAbuGhazal/SurfaceBaisedMLevelContext}{https://github.com/SultanAbuGhazal/SurfaceBaisedMLevelContext}
}

\onecolumn \maketitle \normalsize \setcounter{footnote}{0} \vfill

\section{\uppercase{Introduction}}
\label{sec:introduction}
\vspace{2mm}
Object detection in 3D point clouds, similar to object detection in images, is the task of localizing one or more objects in a 3D scene by means of a 3D bounding box and identifying, or classifying, those objects. In addition to being irregularly spread, point clouds are also unordered datasets; sorting a list of points in a point cloud in different ways does not change what shapes the point cloud represents. Moreover, point clouds are meaningful in neighborhoods of points; single, or isolated, 3D points do not convey any information. Therefore, interactions between neighboring points must be accounted for. Lastly, point cloud representations of shapes change drastically with transformations like shift and rotation, therefore, the learned geometric feature representations must be invariant to such transformations.
\par
Several previous works propose different methods to achieve reliable 3D object detection. Most notably, \cite{dahua2013holistic} propose a method that utilizes geometric features, along with 2D image segmentation of 3D scenes to achieve holistic scene understanding and predict objectness scores and object bounding boxes. This line of methods suffer with occlusion and drastic changes in scanning angle. 
\par
Other detectors use template-based methods where prebuilt databases of common 3D objects are searched using keypoints from a 3D scene to find matches to the objects in the scene. In some cases \cite{yangyan2015dbassisted,liangliang2012searchclassify,litany2017asist}, the matching hand-crafted object models from the prebuilt database are scaled and oriented to match the scanned object and swapped with the object in the scene. \cite{shuran2014sliding} create synthetic depth maps of objects using hundreds of 2D renderings of a single object category to generate point clouds which are used to fit an SVM. During testing, a 3D sliding window is moved across the 3D scene where the overlapped 3D points are classified using the SVM trained earlier at each step. Similar to sliding windows, a follow up work \cite{shuran2016deepsliding} proposes the use of Region Proposal Networks (RPNs) to detect objects in point clouds which is significantly more computationally extensive than RPNs for object detection in 2D images. The more recent \cite{hou2019sis} uses a combination of point clouds and 2D images to train the model on 3D object detection tasks.
\par
As for point-based 3D detectors that consume point clouds with minimal preprocessing, \cite{xie2020mlcvnet} propose a method to capture contextual information from point clouds on multiple levels for 3D object detection tasks with self-attention and multi-scale feature fusion. A 3D detection network, built on the success of \cite{qi2019deep}, that recognizes correlations between 3D objects and the global scene. It introduces three context modules at different stages of \cite{qi2019deep}; at the seed points level, at the vote cluster level, and a module is added to encode a global scene feature. The current best performing 3D detection network by \cite{wang2022rbgnet}, uses an effective sampling and grouping scheme that is biased towards points on objects surfaces. It replaces the sampling and grouping stage of \cite{qi2019deep} with a ray-based grouping stage, and introduces a Foreground Biased Sampling (FBS) scheme to select seed points replacing Furthest Point Sampling (FPS) in the backbone. It uses \cite{qi2017pointnetplusplus} as backbone to extract point features and perform point set abstraction, similar to \cite{qi2019deep}. Additionally, this detector borrows a voting module from \cite{qi2019deep} to generate one vote from each seed point.
\par
In the following sections, the relevant related work is discussed and the different components of the proposed architecture are detailed in sequence. Then, the conducted experiments are presented with a discussion of the obtained results.

\section{\uppercase{Related Work}}
Point clouds are usually a highly irregular data format. The data collection process does not guarantee a uniform distribution of the captured data points. This property makes it challenging to apply common convolution operations in extracting features from 3D point clouds. Traditionally, multiple different methods are devised to work around the irregular nature of point clouds. An intuitive method is using \textit{Volumetric CNNs} \cite{3dshapenet,volumetriccnns,voxnet,vote3d,fpnn,voxel_centerbased,voxel_partaware,voxel_rcnn,voxel_sliding}, or \textit{Voxelization}, where the input point clouds are interpolated onto a 3D voxel grid. In such methods, the resolution of the resulting voxel representation is decided by the voxels' dimensions. Voxels are analogous to 3D pixels, they form a regular data format with uniform density. This representation allows for the use of convolution operations to extract features from the 3D space. Those 3D operations are computation intensive, especially with higher resolution. 
\par
Another method is using \textit{Multi-view CNNs} \cite{multiview3dcnn,shrec16track} where point clouds in 3D are rendered, or projected, on 2D planes which are then put through CNNs for classification. This line of methods perform very well but largely restrict context understanding and do not extend to semantic, per-point, classification. Alternate methods use meshes of 3D data and apply \textit{Spectral CNNs} to classify organic objects with manifold meshes \cite{spectral_scale_invariant,shapenet}. 
\par 
This work aims to extend and further develop point-based 3D detector, due to their simplicity and the relatively less demanding computation requirements. Point-based methods also limit the amount of information loss that can ocurr in voxel-based and projection-based methods.

\subsection{Multi-Level Context VoteNet}
\cite{xie2020mlcvnet} propose MLCVNet, a method to capture contextual information from point clouds on multiple levels for 3D object detection tasks with self-attention and multi-scale feature fusion. Existing 3D object detectors identified objects independently and did not consider the utilization of contextual information between objects in the scene. Meanwhile, MLCVNet recognizes correlations between 3D objects and the global scene. It introduces three context modules at different stages of \cite{qi2019deep} to embed contextual cues on multiple levels. At the seed points level, a Patch-to-Patch Context (PPC) module is introduced to capture relationships between point patches, before the voting stage where the seed points vote for corresponding object centroid points. At the vote cluster level, an Object-to-Object Context (OOC) module is added after the clustering stage, before the prediction stage, to capture correlations between object proposals. Lastly, a third module is added to encode a global scene feature, called the Global Scene Context (GSC) module.

\par
The PPC module aims to capture relationships between point patches in the input point cloud. The point patches in this implementation correspond to the seed points that are generated using the PointNet++ backbone by abstracting the input point set. This module provides two advantages, first, the gathered cross-patch information reduces the effect of missing data in partial patches by supplementing them from other similar patches. Additionally, this module captures cross-patch relationships enhancing the produced votes by combining the voting information of every patch with that of the other patches. This is realized using a Compact Generalized Non-Local, or CGNL, self-attention module \cite{CGNLNetwork2018} where information from all patches are combined channel-wise before the voting stage.

\begin{equation}
    \label{eq:ppc_selfattention}
    \textbf{A}'=f(\theta (\textbf{A}),\phi (\textbf{A}))g(\textbf{A}),
\end{equation}
here, \(\theta(\cdot)\), \(\phi (\cdot)\), \(g(\cdot)\) are transform functions and \(f(\cdot , \cdot)\) is a function for encoding similarities among different locations in the input features.

\par
In \cite{qi2019deep}, vote clusters are processed individually and independently to generate bounding box and class predictions. This does not allow the network to capture relationships among different objects that are often found close to each other. Therefore, the OOC module implements a technique used in 2D object detection for images \cite{Chen2018ContextRF} to allow correlated objects to share weighted information across. Similar to \cite{qi2019deep}, after the votes are clustered around the selected \(K\) vote cluster centers, each cluster is abstracted into a single feature vector, with a location component and a feature component, using an MLP and a max-pooling layer as shown in Eq. \ref{eq:ooc_selfattention}, where \(v_{i}\) is each vote in the cluster and \(n\) is the total count of votes in that cluster. Then, the generated features are put through a self-attention step to capture inter-cluster, or inter-object, relationships before the proposal and classification stage. The CGNL self-attention module generates new cluster features that encode the correlations between all clusters.

\begin{equation}
    \label{eq:ooc_selfattention}
    \mathcal{C}_{OOC}=\textrm{CGNL}\left ( \max_{i=1,\cdots ,n} \{ MLP(v_{i}) \} \right ),
\end{equation}

Finally, the third context module aims to capture the context of the whole scene. This module contributes by encoding global contextual information from the scene into the generated features before the proposal and classification stage. Inspired by the structure inference proposed by \cite{Liu2018StructureIN} for 2D object detection, the GSC module is designed to leverage the global scene context cues to enhance the feature representations of the \(K\) object candidates. This is achieved by adding a branch that connects seed features, or point patches, with object proposals before the PPC and OOC modules.
\par
The GSC module first applies channel-wise max-pooling on both the \(M\) patches and \(K\) clusters, this results in two feature vectors, one representing patches of points and the other representing clusters of votes. The two resulting vectors are concatenated and further aggregated using an MLP that reduces the feature size to match the proposals size. This global feature vector is then expanded to match the number of object proposals \(K\). Finally, the expanded feature is combined with the object proposals using element-wise multiplication. This operation is expressed in Eq. \ref{eq:gsc_context}.

\begin{equation}
    \label{eq:gsc_context}
    \textbf{\(\mathcal{C}\)}_{new}=MLP([max(\textbf{\(\mathcal{C}\)});max(\textbf{P})])+\textbf{\(\mathcal{C}\)}_{OOC},
\end{equation}

The result of adding the PPC, OOC, then the GSC modules in order is an increase in the baseline mean average precision (mAP) by 2.6, 1.2, and 1.1 percentages, respectively, when trained on detection of 3D objects from the ScanNet dataset. Overall, MLCVNet achieves 64.5 mAP in detection of 3D objects from the ScanNet dataset with an IoU threshold of 0.25.

\subsection{Ray-based Grouping}
\cite{wang2022rbgnet} introduces RBGNet, an effective sampling and grouping scheme that is biased towards points on objects surfaces. In a nutshell, it replaces the sampling and grouping stage in \cite{qi2019deep} with a ray-based grouping stage, and introduces a Foreground Biased Sampling (FBS) scheme to select seed points replacing Furthest Point Sampling (FPS) in the backbone. While the voting scheme adopted by \cite{qi2019deep} plays a significant role in generating object proposals, the clustering and grouping of votes and their corresponding feature vectors is rather generic and simple, which leaves room for improvement. The quality of object proposals is directly affected by that of the grouping and clustering process. Therefore, RBGNet aims to improve the grouping and clustering of votes by proposing a ray-based grouping method that selectively encodes object shape information into the object proposals for better accuracy. Instead of simply clustering votes by proximity to vote cluster centers and directly aggregating features with a set abstraction module, votes are clustered along rays emitted from each vote cluster center and then aggregated in a selective manner taking into consideration the relevance of each vote to the object it belongs to and the object surface geometry. Tthe ray-based grouping module starts with \(M\) cluster centers \(\{c_{i}\} _{i=1}^{M}\), where \(c_{i}=[v_{i};f_{i}]\) represents the vote center position \(v_{i} \in \mathbb{R}^{3}\) and the vote center feature vector \(f_{i} \in \mathbb{R}^{3}\). It then generates \(N\) rays emitting out of the cluster center with angles and lengths standardized based on three rules using spherical coordinates. First, \(P\) polar angles with the z-axis are determined using Eq. \ref{eq:rbgnet_polar_angles}, where \(\theta _{p}\) is the \(p^{th}\) polar angle.

\begin{equation}
    \label{eq:rbgnet_polar_angles}
    \theta _{p}=\frac{\pi p}{(P-1)},\; p \in \{0,\cdots,P-1\}
\end{equation}

Second, the \(N\) rays to be generated are distributed across the \(P\) polar angles following the cases in Eq. \ref{eq:rbgnet_ray_distribution}. This always results in exactly one ray pointing in the positive z-axis direction and exactly one ray pointing in the negative z-axis direction, and \(A_{p}\) rays on the \(p^{th}\) polar angle. in Eq. \ref{eq:rbgnet_ray_distribution} the number 4 is a hyper-parameter that can be modified along with the total number of rays \(N\) and the number of polar angles \(P\).

\begin{equation}
    \label{eq:rbgnet_ray_distribution}
    A_{p}=\begin{cases}
    1, & \text{ if } \; p=0 \; \textup{or} \;  P-1 \\ 
    4\times p, & \text{ if } \; 0< p\leq \frac{P-1}{2} \\ 
    4\times (P-p-1), & \text{ if } \; P-1>p>\frac{P-1}{2}
    \end{cases}
\end{equation}

Third, the azimuth angles of the rays on each polar angle are determined in such a way that the \(A_{p}\) rays on the \(p^{th}\) polar angle are evenly spaced around the azimuth of \(2\pi\) forming rings perpendicular to the z-axis. Both the polar \(\theta _{p,a} \in [0, \pi]\) and azimuth \(\psi _{p,a} \in [0, 2\pi]\) angles of each ray is determined using Eq. \ref{eq:rbgnet_ray_angles}.

\begin{equation}
    \label{eq:rbgnet_ray_angles}
    \psi _{p,a}=\frac{2 \pi a}{A_{p}}, \;\; \theta_{p,a}=\theta _{p}, \;\; a \in \{0, \cdots, A_{p}-1\}
\end{equation}

\par
Finally, to determine how far the rays extend away from the vote cluster center, the module learns to predict each object's scale \(l_{i}\), that represents half length of the diagonal side of the object bounding box, using the cluster feature vector \(f_{i}\). This sub-network is supervised using smooth-\(\ell _{1}\) norm as shown in Eq. \ref{eq:rbgnet_object_scale_loss} where \(l_{i}^{*}\) is half length of the diagonal side of the object's ground-truth bounding box, and \(I\) is the count of positive cluster centers. The indicator function \(\mathbb{I}[i_{th} \textup{ is positive}]\) indicates if the cluster center \(c_{i}\) is within a 0.3 meters of the ground-truth object center.

\begin{equation}
    \label{eq:rbgnet_object_scale_loss}
    L_{scale-reg}=\frac{1}{I}\sum_{i} \left \| l_{i} -l_{i}^{*} \right \|_{\eta } \mathbb{I}[i_{th} \textup{ is positive}]
\end{equation}

\par
Determining the rays to generate is only the first step in the ray-based grouping module. To realize the benefits of focusing on objects surface geometry and fine-grained shape features, points along each ray are selected in a systematic, coarse-to-fine, way for feature extraction, following \cite{mildenhall2020nerf}. Essentially, the length of each ray is divided into sections where each section is sampled with two layers of detail, coarse and fine, depending on the position of the seed points in each section and their density. In each layer, anchor points are generated and seed points are systematically sampled around those anchor points. Prior to the coarse and fine feature extraction processes, the seed points are upsampled from 1024 points back to 2048 points using information from the first set abstraction layer in the backbone.

\par
Starting with the coarse sampling process, \(K_{c}\) anchor points are generated along each of the \(N\) rays with equal spacing dividing the ray into \(K_{c}\) sections, or bins. This set of anchor points can be represented as in Eq. \ref{eq:rbgnet_coarse_anchors} where \(n\) denotes the \(n^{th}\) ray. Then, seed points are sampled around each anchor point \(q _{n,k}^{(c)}\) depending on their proximity, and processed using a set abstraction layer to generate a single local feature \(\rho _{n,k}^{(c)}\) representing the corresponding section of the ray.

\begin{equation}
    \label{eq:rbgnet_coarse_anchors}
    Q_{n}^{(c)}=\{q_{n,k}^{(c)}=(x_{n,k}^{(c)}, y_{n,k}^{(c)}, z_{n,k}^{(c)})\}, \;\; k \in \{1,\cdots ,K_{c} \}
\end{equation}

To selectively weigh the anchor points, a classification sub-network is trained to give each anchor point a positive label if it lies on the object surface and a negative label otherwise. This enables predicting which anchor points, and their surrounding seed points, belong to the object based on the local feature vector of the anchor points \(\rho _{n,k}^{(c)}\) and the vote cluster feature vector \(f_{i}\) as input. The predicted surface mask of positive anchor points \(m_{n,k}^{(c)}\) is represented by Eq. \ref{eq:rbgnet_anchor_mask}. Training this sub-network is supervised by giving a positive label to anchor points that have some points from the ground-truth object surface within its proximity, and a negative label otherwise.
\par
This process is repeated when generating fine anchor points. The result is the coarse local features set \(\mathcal{P}^{(c)}=\{\rho _{n,k}^{(c)}\}_{K_{c}, N}^{k=1,n=1}\), the fine features set \(\mathcal{P}^{(f)}=\{\rho _{n,k}^{(f)}\}_{K_{f}, N}^{k=1,n=1}\), the set of surface seed points masks for coarse points \(\mathcal{M}^{(c)}=\{m^{(c)}_{n,k}\}_{K_{c}, N}^{k=1,n=1}\) and for fine points \(\mathcal{M}^{(f)}=\{m^{(f)}_{n,k}\}_{K_{f}, N}^{k=1,n=1}\), and the locations of the corresponding generated anchor points.

\begin{equation}
    \label{eq:rbgnet_anchor_mask}
    m_{n,k}^{(c)}=\mathcal{F}_{mask}^{(c)}(\rho _{n,k}^{(c)}, f_{i})
\end{equation}

\par
Finally, the extracted features are combined in a fusion stage. The coarse and fine surface point masks are first applied to the coarse and fine features, generating the masked feature sets \(\hat{\mathcal{P}}^{(c)}\) and \(\hat{\mathcal{P}}^{(f)}\), where the features corresponding to a negative anchor point in the mask is set to zero. Then, the coarse and fine features are put through two separated aggregation branches. The masked coarse point features of a given ray \(n\), \(\{\hat{\rho} _{n,k}^{c}\}_{k=1}^{K_{c}}\) are concatenated in order into a single feature \(r_{n}^{(c)}\) representing the ray and the projected to a feature vector of size 32. This is represented in Eq. \ref{eq:rbgnet_point_fusion} where the concatenation is denoted by \(\odot\).

\begin{equation}
    \label{eq:rbgnet_point_fusion}
    r_{n}^{(c)}=\mathcal{F}_{\textrm{point}}^{(c)}(\{\hat{\rho} _{n,k}^{(c)}\} _{k=1}^{K_{c}},\odot )
\end{equation}

Similarly, the coarse ray features \(\mathcal{R}^{(c)} = \{r_{n}^{(c)}\} _{n=1}^{N}\) are concatenated in order and processed with an MLP of two hidden layers and an output size of 128, represented in Eq. \ref{eq:rbgnet_ray_fusion}. In both concatenation steps, the order of the rays is the same across object proposals.

\begin{equation}
    \label{eq:rbgnet_ray_fusion}
    \mu^{(c)}=\mathcal{F}_{\textrm{ray}}^{(c)}(\{r _{n}^{(c)}\} _{n=1}^{N},\odot )
\end{equation}

As for the fine point feature aggregation branch, the same strategy is followed to obtain a 128-dim feature vector \(\mu ^{(f)}\). At the end of the fusion stage the coarse and fine aggregated vectors are fused, as in Eq. \ref{eq:rbgnet_fusion}, generating the fused feature \(g\) which is in turn combined with the vote cluster feature \(f\) to ultimately enhance the performance of the network.

\begin{equation}
    \label{eq:rbgnet_fusion}
    g=\mathcal{F}_{\textrm{fuse}}(\mu ^{(c)}, \mu ^{(f)})
\end{equation}

\par
RBGNet also proposes a more effective seed point sampling method to replace FPS in the backbone \cite{qi2017pointnetplusplus}. The FPS algorithm usually ends up heavily sampling background points that do not belong to objects of interest, specifically in the first set abstraction layer that downsamples the points to 2048 points. Therefore, FBS separately applies FPS on foreground points and background points. To distinguish the foreground from background points, an additional segmentation head is added to the backbone to predict foreground points supervised by the masks provided by the ground-truth bounding boxes. The segmentation head predicts scores for each point with \(\varrho _{j}\) location and \(\nu _{j}\) feature vector to be on the foreground or not as shown in Eq. \ref{eq:rbgnet_fore}.

\begin{equation}
    \label{eq:rbgnet_fore}
    \varepsilon _{j}=\mathcal{F}^{\textrm{fore}}(\nu _{j}, \varrho _{j}) \in [0,1]
\end{equation}

The points are then sorted by score where the top \(\kappa\) points are labeled as foreground points \(\mathcal{D}^{(f)}=\{d_{j}^{(f)}\}_{j=1}^{\kappa}\) and the rest are labeled as background points \(\mathcal{D}^{(b)}=\{d_{j}^{(b)}\}_{j=1}^{2048 - \kappa}\). Finally, FPS is applied on both sets of points and the results are combined in one sample set as in Eq. \ref{eq:rbgnet_combined_set}, where \(\mathcal{S}\) is the final combined set of points.

\begin{multline}
    \label{eq:rbgnet_combined_set}
    \mathcal{D}^{\hat{(f)}}=\textup{FPS}(\mathcal{D}^{(f)}), \\ 
    \mathcal{D}^{\hat{(b)}}=\textup{FPS}(\mathcal{D}^{(b)}), \;\; \mathcal{S}=\mathcal{D}^{\hat{(f)}} \oplus \mathcal{D}^{\hat{(b)}}
\end{multline}

Introducing ray-based grouping module and FBS pushes the accuracy of the 3D object detection network surpassing contemporary models \cite{Liu2021GroupFree3D,Zhang2020H3DNet}. RBGNet achieves a maximum mean average precision (mAP) of 70.6 in detection of 3D objects from the ScanNet dataset with an IoU threshold of 0.25 over 25 runs averaging at 69.9.

\section{\uppercase{Method}}
\begin{figure*}[!h]
  \centering
   {\epsfig{file = 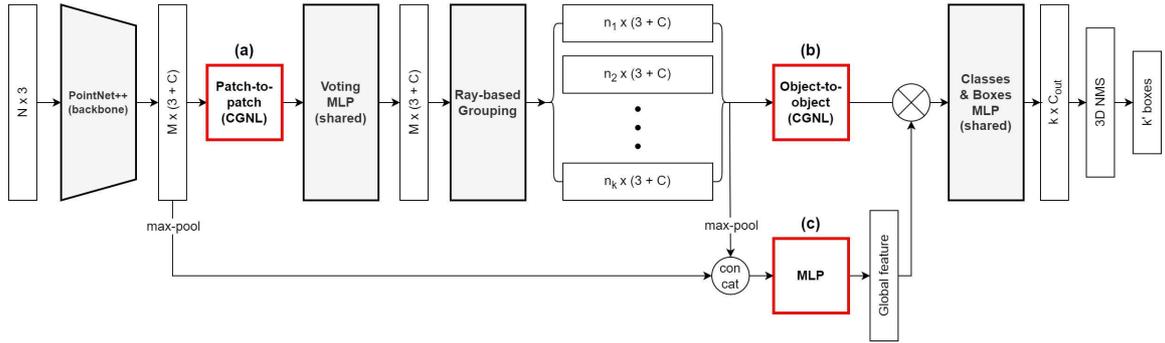, width = 0.96\textwidth}}
  \vspace{-1mm}
  \caption{The proposed architecture incorporating all CGNL self-attention modules and the global feature, (a) the patch-to-patch context module to encode similar point patches, (b) the object-to-object context module to encode correlated object candidates, and (c) the global scene context module to aggregate low and high level features.}
  \vspace{-2mm}
  \label{fig:new_architecture}
\end{figure*}

The proposed architecture incorporates a patch-level self-attention module, an object-level self-attention module, and a global context feature along with ray-based coarse-to-fine feature grouping to generate object candidates. As shown in Fig. \ref{fig:new_architecture}, the input point cloud first passes through a PointNet++ backbone for feature extraction and point set abstraction, which reduces the \(N\) input points to \(M\) seed points. The set abstraction modules in PointNet++ generally start with a \(N \times (d+C)\) input matrix, where \(N\) is the number of points, \(d\) is the dimension of location vector and \(C\) is the dimension of the feature vector. This input is first sampled using FPS for the best coverage of the \(M\) sampled seed points. The input point cloud is then divided into groups, or local regions, around the \(M\) seed points, making the output \(M \times n \times (d+C)\) where \(n\) is the group size. The group size can vary across groups, but fortunately, the PointNet \cite{pointnet} module can map any number of points to a fixed feature vector. Finaly, this output is abstracted using a PointNet layer generating the output \(M \times (d+C')\).

This is followed by a Compact Generalized Non-Local (CGNL) self-attention module that encodes patch-to-patch information, Fig. \ref{fig:new_architecture}a. This enhances the patch features by incorporating information across similar patches. The CGNL module leverages the channel-wise correlations in the point features as well as the relations between every point patch pair.

The seed point features, with the self-attention information encoded, are then passed to the voting module. The voting MLP, predicts one vote towards the object center for each seed point \(s_{i}=\left [ x_{i};f_{i} \right ]\) knowing \(x_{i}\in \mathbb{R}^{3}\), and \(f_{i}\in \mathbb{R}^{C}\) \cite{qi2019deep}. This module predicts an offset in Euclidean space \(\Delta x_{i}\in \mathbb{R}^{3}\) as well as an offset in feature space \(\Delta f_{i}\in \mathbb{R}^{C}\) making each vote for each seed \(s_{i}\) representable as \(v_{i}=\left [ y_{i};g_{i} \right ]\) where \(y_{i}=x_{i}+\Delta x_{i}\) and \(g_{i}=f_{i}+\Delta f_{i}\). The ground-truth bounding box centers are used in training to regularize the object center offset learning using Eq. \ref{eq:voting_loss}.

\begin{multline}
    \label{eq:voting_loss}
    L_{vote-reg}= \\ 
    \frac{1}{M_{pos}}\sum_{i} \left \| \Delta x_{i}-\Delta x_{i}^{*} \right \| \mathbbm{1} \left [ s_{i} \textrm{ on object} \right ],
\end{multline}
where \(\mathbbm{1} \left [ s_{i} \textrm{ on object} \right ] \) is an indicator function that indicates if the seed point is on the surface of the object or not, \( M_{pos} \) is the total number of seeds on the surface of the object, and \( \Delta x_{i}^{*}\) is the offset of the seed point from the corresponding object's ground-truth bounding box center.

The overall learning objective can be represented by Eq. \ref{eq:overall_loss}, where \(\mathcal{L}_{\textup{fbs}}\) is the foreground-biased sampling loss, \(\mathcal{L}_{\textup{rbfg}}\) is the ray-based grouping loss, \(\mathcal{L}_{\textup{obj-cls}}\) is the objectness score loss, \(\mathcal{L}_{\textup{box}}\) is the bounding box prediction loss, and \(\mathcal{L}_{\textup{sem-cls}}\) is the semantic classification loss.

\begin{multline}
    \label{eq:overall_loss}
    L=\lambda _{\textup{vote-reg}}\mathcal{L}_{\textup{vote-reg}}+\lambda _{\textup{fbs}}\mathcal{L}_{\textup{fbs}}+\lambda _{\textup{rbfg}}\mathcal{L}_{\textup{rbfg}} \\ 
    +\lambda _{\textup{obj-cls}}\mathcal{L}_{\textup{obj-cls}}+\lambda _{\textup{box}}\mathcal{L}_{\textup{box}} \\ 
    +\lambda _{\textup{sem-cls}}\mathcal{L}_{\textup{sem-cls}}
\end{multline}

\par
The resulting votes are clustered and grouped by the ray-based grouping module \cite{wang2022rbgnet} which produces \(K\) clusters, or object candidates. This module performs coarse-to-fine feature extraction on the vote centers, and captures the objects surface geometry to enhance the grouping of vote centers into object candidates. It first upsamples the points back to 2048 points, then samples them in two levels, coarse and fine, along 18 generated rays and passes them through set abstraction layers. The extracted features are fused systematically to obtain a single feature vector for each object candidate.

To regularize the ray-based feature grouping, the \(\mathcal{L}_{\textup{rbfg}}\) loss is defined as the sum of coarse and fine cross entropy losses, \(\mathcal{L}_{\textup{c-cls}}\) and \(\mathcal{L}_{\textup{f-cls}}\), and the object scale smooth-\(\ell _{1}\) loss \(\mathcal{L}_{\textup{scale-reg}}\), as shown in Eq. \ref{eq:rbfg_loss}.

\vspace{-3mm}
\begin{multline}
    \label{eq:rbfg_loss}
    \mathcal{L}_{\textup{rbfg}}=\lambda _{\textup{scale-reg}}\mathcal{L}_{\textup{scale-reg}} \\ 
    +\lambda _{\textup{c-cls}}\mathcal{L}_{\textup{c-cls}}+\lambda _{\textup{f-cls}}\mathcal{L}_{\textup{f-cls}}
\end{multline}

\par
The object candidates are then processed with the second CGNL self-attention module to encode object level correlations, Fig. \ref{fig:new_architecture}b. At this point, the output of this module is combined with a global feature that is obtained by aggregating patch and cluster features, Fig. \ref{fig:new_architecture}c. The combined feature vectors are finally passed through a fully connected layer for classification and bounding box prediction.

\begin{table}[h]
\vspace{2mm}
\caption{The RBGNet baseline results, obtained by 5 training runs on the ScanNet dataset. For Evalutation, IoU threshold is at 0.25. The maximum and average mAP@0.25 are lower than the numbers reported in \cite{wang2022rbgnet} where the reported maximum mAP is 70.6 and 69.9 on average after performing 25 training runs on ScanNet.}\label{tab:runs_baseline} \centering
\resizebox{0.475\textwidth}{!}{%
\begin{tabular}{l|lllll|ll}
\hline
\textbf{}       & \textbf{R1} & \textbf{R2} & \textbf{R3} & \textbf{R4} & \textbf{R5} & \textbf{max}  & \textbf{avg} \\ \hline
\textbf{mAP}    & 68.4        & 69.2        & 69.6        & 69.5        & 69.1        & \textbf{69.6} & 69.2         \\ \hline
\textbf{cbnt}   & 52.2        & 54.9        & 53.4        & 53.8        & 51.5        & 54.9          & 53.1         \\
\textbf{bed}    & 89.7        & 91.1        & 87.1        & 91.9        & 88.8        & 91.9          & 89.7         \\
\textbf{chair}  & 94.0        & 93.9        & 93.8        & 93.0        & 94.1        & 94.1          & 93.8         \\
\textbf{sofa}   & 91.3        & 91.0        & 90.6        & 91.6        & 92.5        & 92.5          & 91.4         \\
\textbf{table}  & 73.9        & 73.7        & 74.2        & 75.6        & 74.5        & 75.6          & 74.4         \\
\textbf{door}   & 61.3        & 62.5        & 60.0        & 63.0        & 60.9        & 63.0          & 61.5         \\
\textbf{wdw}    & 56.4        & 53.0        & 55.1        & 57.2        & 57.9        & 57.9          & 55.9         \\
\textbf{bkslf}  & 55.4        & 62.5        & 61.8        & 58.8        & 61.0        & 62.5          & 59.9         \\
\textbf{pic}    & 22.8        & 23.2        & 21.3        & 20.8        & 23.6        & 23.6          & 22.3         \\
\textbf{cntr}   & 67.0        & 65.8        & 75.3        & 74.9        & 63.3        & 75.3          & 69.3         \\
\textbf{desk}   & 76.7        & 76.2        & 76.7        & 78.3        & 79.5        & 79.5          & 77.5         \\
\textbf{crtn}   & 53.0        & 49.1        & 51.7        & 49.3        & 52.2        & 53.0          & 51.0         \\
\textbf{refrg}  & 50.6        & 50.4        & 58.9        & 54.6        & 51.0        & 58.9          & 53.1         \\
\textbf{shcrtn} & 73.9        & 72.6        & 74.0        & 70.2        & 75.4        & 75.4          & 73.2         \\
\textbf{toilet} & 95.3        & 99.8        & 98.8        & 99.6        & 99.1        & 99.8          & 98.5         \\
\textbf{sink}   & 67.4        & 73.7        & 68.6        & 68.3        & 65.8        & 73.7          & 68.7         \\
\textbf{tub}    & 92.7        & 94.3        & 92.1        & 93.6        & 95.0        & 95.0          & 93.5         \\
\textbf{gbin}   & 58.1        & 57.1        & 60.1        & 56.0        & 57.5        & 60.1          & 57.8         \\ \hline
\end{tabular}}
\end{table}

\subsection{Dataset and Evaluation Metric}
The proposed architecture is evaluated on ScanNet \cite{scannet2017} a large indoor 3D scene dataset following the standard training and evaluation splits. The dataset does not include ground-truth bounding boxes by default, therefore, the boxes are generated from the provided semantic point labels. ScanNet consists of 1513 samples with labels for 18 different object classes. As for the evaluation metric, mean average precision (mAP) is used with a 0.25 IoU threshold following \cite{qi2019deep}.

\vspace{-4mm}
\section{\uppercase{Results}}
For the experiments baseline, the RBGNet \cite{wang2022rbgnet} network is trained on the ScanNet dataset as recommended by the authors using 4 GPUs and with identical configuration. Table \ref{tab:runs_baseline} lists the obtained baseline results.

\subsection{Patch Attention}
In the first experiment, only the patch-to-patch context module is appended to the network. This is done by reshaping the output of the feature extraction backbone and passing them through a CGNL module. Table \ref{tab:runs_exp1} lists the results obtained by training the network on the ScanNet dataset. It is observed that this simple modification has the most significant contribution to higher overall performance.

\begin{table}[h]
\caption{The results after incorporating the CGNL self-attention module on the patch level (Patch-to-Patch Context), obtained by 5 training runs on the ScanNet dataset. For Evalutation, IoU threshold is at 0.25.} \label{tab:runs_exp1} \centering
\resizebox{0.475\textwidth}{!}{%
\begin{tabular}{l|lllll|ll}
\hline
\textbf{}       & \textbf{R1} & \textbf{R2} & \textbf{R3} & \textbf{R4} & \textbf{R5} & \textbf{max}  & \textbf{avg} \\ \hline
\textbf{mAP}    & 70.4        & 69.2        & 70.5        & 69.3        & 69.3        & \textbf{70.5} & 69.7         \\ \hline
\textbf{cbnt}   & 54.3        & 55.4        & 53.5        & 56.8        & 51.2        & 56.8          & 54.2         \\
\textbf{bed}    & 90.2        & 90.3        & 92.0        & 91.1        & 91.6        & 92.0          & 91.0         \\
\textbf{chair}  & 94.0        & 95.0        & 94.2        & 94.9        & 93.8        & 95.0          & 94.4         \\
\textbf{sofa}   & 92.8        & 91.4        & 91.7        & 90.8        & 91.0        & 92.8          & 91.5         \\
\textbf{table}  & 75.1        & 75.6        & 76.2        & 74.0        & 74.2        & 76.2          & 75.0         \\
\textbf{door}   & 64.5        & 62.3        & 61.6        & 61.2        & 62.1        & 64.5          & 62.3         \\
\textbf{wdw}    & 58.6        & 55.3        & 58.0        & 53.9        & 56.7        & 58.6          & 56.5         \\
\textbf{bkslf}  & 58.5        & 56.4        & 65.5        & 49.7        & 54.2        & 65.5          & 56.9         \\
\textbf{pic}    & 25.8        & 22.3        & 21.8        & 21.4        & 21.6        & 25.8          & 22.6         \\
\textbf{cntr}   & 68.9        & 62.3        & 64.0        & 66.6        & 70.5        & 70.5          & 66.5         \\
\textbf{desk}   & 77.6        & 78.0        & 79.1        & 77.3        & 76.3        & 79.1          & 77.7         \\
\textbf{crtn}   & 56.9        & 53.0        & 50.9        & 55.2        & 54.3        & 56.9          & 54.1         \\
\textbf{refrg}  & 51.5        & 54.9        & 61.1        & 59.0        & 58.5        & 61.1          & 57.0         \\
\textbf{shcrtn} & 73.8        & 71.9        & 78.1        & 78.0        & 74.6        & 78.1          & 75.3         \\
\textbf{toilet} & 99.1        & 97.3        & 97.4        & 98.6        & 97.3        & 99.1          & 97.9         \\
\textbf{sink}   & 71.0        & 71.9        & 71.4        & 66.9        & 67.2        & 71.9          & 69.7         \\
\textbf{tub}    & 95.6        & 93.3        & 94.2        & 94.0        & 93.8        & 95.6          & 94.2         \\
\textbf{gbin}   & 58.7        & 58.9        & 57.9        & 57.2        & 58.0        & 58.9          & 58.1         \\ \hline
\end{tabular}}
\end{table}

\subsection{Patch and Object Attention}
In the second experiment, the patch-to-patch context module is accompanied by the object-to-object context module where both are appended to the network. To accommodate the new module, the cluster features are reshaped and passed through two consecutive CGNL modules. Table \ref{tab:runs_exp3} lists the results obtained by training the network on the ScanNet dataset. In the presented experiments, this combination does not perform as well as the patch-to-patch experiment and stay on par with the baseline.

\begin{table}[h]
\caption{The results after incorporating the CGNL self-attention module on the patch level (Patch-to-Patch Context) and on the object candidates level (Object-to-Object Context), obtained by 5 training runs on the ScanNet dataset. For Evalutation, IoU threshold is at 0.25.} \label{tab:runs_exp3} \centering
\resizebox{0.475\textwidth}{!}{%
\begin{tabular}{l|lllll|ll}
\hline
\textbf{}       & \textbf{R1} & \textbf{R2} & \textbf{R3} & \textbf{R4} & \textbf{R5} & \textbf{max}  & \textbf{avg} \\ \hline
\textbf{mAP}    & 69.7        & 69.8        & 69.4        & 69.6        & 66.8        & \textbf{69.8} & 69.1         \\ \hline
\textbf{cbnt}   & 55.7        & 52.1        & 53.8        & 55.3        & 49.5        & 55.7          & 53.3         \\
\textbf{bed}    & 89.7        & 89.4        & 88.7        & 89.6        & 89.5        & 89.7          & 89.4         \\
\textbf{chair}  & 94.0        & 94.2        & 93.4        & 94.0        & 93.5        & 94.2          & 93.8         \\
\textbf{sofa}   & 91.6        & 88.9        & 91.7        & 90.0        & 87.2        & 91.7          & 89.9         \\
\textbf{table}  & 74.3        & 75.1        & 76.9        & 74.2        & 72.6        & 76.9          & 74.6         \\
\textbf{door}   & 61.6        & 62.2        & 60.4        & 62.4        & 57.6        & 62.4          & 60.8         \\
\textbf{wdw}    & 57.8        & 57.4        & 56.3        & 54.5        & 58.0        & 58.0          & 56.8         \\
\textbf{bkslf}  & 60.7        & 60.7        & 62.1        & 60.0        & 56.5        & 62.1          & 60.0         \\
\textbf{pic}    & 22.5        & 22.2        & 23.3        & 23.6        & 19.1        & 23.6          & 22.1         \\
\textbf{cntr}   & 65.2        & 67.1        & 66.8        & 68.3        & 48.0        & 68.3          & 63.1         \\
\textbf{desk}   & 79.0        & 78.5        & 77.0        & 82.2        & 74.1        & 82.2          & 78.1         \\
\textbf{crtn}   & 54.1        & 56.9        & 54.8        & 52.3        & 60.0        & 60.0          & 55.6         \\
\textbf{refrg}  & 55.1        & 60.8        & 55.5        & 52.6        & 48.3        & 60.8          & 54.5         \\
\textbf{shcrtn} & 74.5        & 70.3        & 71.3        & 76.6        & 72.5        & 76.6          & 73.0         \\
\textbf{toilet} & 99.4        & 99.6        & 99.9        & 98.8        & 97.3        & 99.9          & 99.0         \\
\textbf{sink}   & 72.2        & 72.2        & 66.1        & 67.6        & 71.5        & 72.2          & 69.9         \\
\textbf{tub}    & 91.7        & 91.7        & 92.5        & 94.9        & 89.8        & 94.9          & 92.1         \\
\textbf{gbin}   & 56.1        & 57.5        & 58.2        & 56.7        & 56.5        & 58.2          & 57.0         \\ \hline
\end{tabular}}
\end{table}

\vspace{-3mm}
\subsection{Patch and Object Attention with Global Feature}
In the final experiment, the patch-to-patch and the object-to-object context modules are kept as is and a global feature is introduce to the network. To generate the global feature, both the seed features and cluster features are max-pooled and concatenated then aggregated using an MLP to reduce the feature size to match that of the cluster features. Table \ref{tab:runs_exp4} lists the results obtained by training the network on the ScanNet dataset. This combination out performs the baseline and achieves higher detection performance. The maximum mAP achieved in 5 training runs is 70.2 averaging at 69.6. On the other hand, the baseline achieves a maximum mAP of 69.6 averaging at 69.2.

\begin{table}[h]
\caption{The results after incorporating the CGNL self-attention module on the patch level (Patch-to-Patch Context) and on the object candidates level (Object-to-Object Context) along with the global context feature, obtained by 5 training runs on the ScanNet dataset. For Evalutation, IoU threshold is at 0.25.} \label{tab:runs_exp4} \centering
\resizebox{0.475\textwidth}{!}{%
\begin{tabular}{l|lllll|ll}
\hline
\textbf{}       & \textbf{R1} & \textbf{R2} & \textbf{R3} & \textbf{R4} & \textbf{R5} & \textbf{max}  & \textbf{avg} \\ \hline
\textbf{mAP}    & 69.2        & 70.2        & 69.8        & 69.8        & 69.2        & \textbf{70.2} & 69.6         \\ \hline
\textbf{cbnt}   & 54.4        & 57.6        & 55.3        & 55.8        & 52.1        & 57.6          & 55.0         \\
\textbf{bed}    & 90.4        & 91.4        & 88.7        & 90.9        & 90.4        & 91.4          & 90.3         \\
\textbf{chair}  & 94.3        & 93.8        & 93.7        & 93.4        & 93.7        & 94.3          & 93.8         \\
\textbf{sofa}   & 90.4        & 88.8        & 91.9        & 92.2        & 90.7        & 92.2          & 90.8         \\
\textbf{table}  & 75.1        & 74.6        & 73.6        & 73.2        & 75.8        & 75.8          & 74.4         \\
\textbf{door}   & 61.5        & 59.9        & 61.5        & 62.5        & 62.4        & 62.5          & 61.5         \\
\textbf{wdw}    & 53.4        & 57.0        & 57.1        & 57.9        & 57.7        & 57.9          & 56.6         \\
\textbf{bkslf}  & 58.6        & 56.4        & 57.0        & 55.6        & 61.2        & 61.2          & 57.8         \\
\textbf{pic}    & 23.6        & 24.7        & 22.6        & 22.9        & 24.1        & 24.7          & 23.6         \\
\textbf{cntr}   & 65.1        & 67.3        & 62.9        & 69.3        & 68.7        & 69.3          & 66.7         \\
\textbf{desk}   & 78.9        & 80.3        & 81.4        & 82.2        & 79.9        & 82.2          & 80.5         \\
\textbf{crtn}   & 53.9        & 58.9        & 49.6        & 54.7        & 52.0        & 58.9          & 53.8         \\
\textbf{refrg}  & 57.1        & 59.3        & 65.3        & 59.2        & 60.2        & 65.3          & 60.2         \\
\textbf{shcrtn} & 71.1        & 73.6        & 75.2        & 73.8        & 69.6        & 75.2          & 72.7         \\
\textbf{toilet} & 99.2        & 98.9        & 97.5        & 99.3        & 99.3        & 99.3          & 98.9         \\
\textbf{sink}   & 67.1        & 70.8        & 71.5        & 67.4        & 68.9        & 71.5          & 69.1         \\
\textbf{tub}    & 92.9        & 92.4        & 93.7        & 89.8        & 84.5        & 93.7          & 90.6         \\
\textbf{gbin}   & 58.9        & 57.6        & 57.3        & 56.4        & 55.1        & 58.9          & 57.0         \\ \hline
\end{tabular}}
\end{table}

\section{\uppercase{Conclusions}}
\label{sec:conclusion}
Embedding contextual cues through the application of self-attention on multiple levels in 3D points is a powerful tool that can significantly enhance feature representation and ultimately 3D scene understanding. Through the experiments presented in this work, self-attention modules are introduced at different levels in a network with accurate ray-based, surface-biased, feature grouping. First at the patch level, then at the object level, and lastly, on the global scene level. The effect of each module was reported through several runs and it was observed that patch level self-attention can contribute significantly to the performance of the 3D object detector. With more intricate integration and further experimentation, the baseline can better leverage the appended self-attention modules, especially, to work better with the fusion stage of the fine and coarse features extracted from the vote clusters.


\bibliographystyle{apalike}
{\small
\bibliography{example}}



\end{document}